\pdfoutput=1

\documentclass[11pt]{article}

\usepackage[preprint]{acl}

\usepackage{times}
\usepackage{latexsym}

\usepackage[T1]{fontenc}

\usepackage[utf8]{inputenc}

\usepackage{microtype}

\usepackage{inconsolata}

\usepackage{graphicx}
\usepackage{amsmath}
\usepackage{multirow}
\usepackage{booktabs}
\usepackage{listings}
\title{PolicyEvol-Agent: Evolving Policy via Environment Perception and Self-Awareness with Theory of Mind}

\author{Yajie Yu, Yue Feng \\
  School of Computer Science\\
  University of Birmingham\\
  United Kingdom\\
  \texttt{yajieyu46@gmail.com}, \texttt{y.feng.6@bham.ac.uk} \\}

\begin{document}
\maketitle
\begin{abstract}
Multi-agents has exhibited significant intelligence in real-word simulations with Large language models (LLMs) due to the capabilities of social cognition and knowledge retrieval. However, existing research on agents equipped with effective cognition chains including reasoning, planning, decision-making and reflecting remains limited, especially in the dynamically interactive scenarios. In addition, unlike human, prompt-based responses face challenges in psychological state perception and empirical calibration during uncertain gaming process, which can inevitably lead to cognition bias. In light of above, we introduce \textbf{PolicyEvol-Agent}, a comprehensive LLM-empowered framework characterized by systematically acquiring intentions of others and adaptively optimizing irrational strategies for continual enhancement. Specifically, \textbf{PolicyEvol-Agent} first obtains reflective expertise patterns and then integrates a range of cognitive operations with Theory of Mind alongside internal and external perspectives. Simulation results, outperforming RL-based models and agent-based methods, demonstrate the superiority of \textbf{PolicyEvol-Agent} for final gaming victory. Moreover, the policy evolution mechanism reveals the effectiveness of dynamic guideline adjustments in both  automatic and human evaluation. \footnote{Our code is released at \url{https://github.com/GuanNiPiShi123/PolicyEvol-Agent}.}
\end{abstract}

\section{Introduction}
\label{Introduction}
Recent achievements based on Large Language Models (LLMs) has demonstrated remarkable potential in problem-solving and real-word simulation tasks due to capabilities of language understanding and text generation \citep{10.5555/3648699.3648939,10.5555/3666122.3668386, zeng2023glmb,Touvron2023LLaMAOA,Achiam2023GPT4TR,wu2024autogen,deepseekai2025deepseekr1incentivizingreasoningcapability}. Therefore, implementing an autonomous LLM-based agent with advanced skills (\emph{e.g.}, reasoning, planning, decision-making, reflecting, and memorizing) in an even more complicated environment has been a sustained attention \citep{yao2023react,10.5555/3600270.3602070,schick2023toolformer,shen2024hugginggpt,hong2024metagpt}. These abilities typically aligns with the expectations of humans for perceiving the surroundings and taking actions in response to complicated requirements.
\begin{figure}[t]
  \includegraphics[width=\columnwidth]{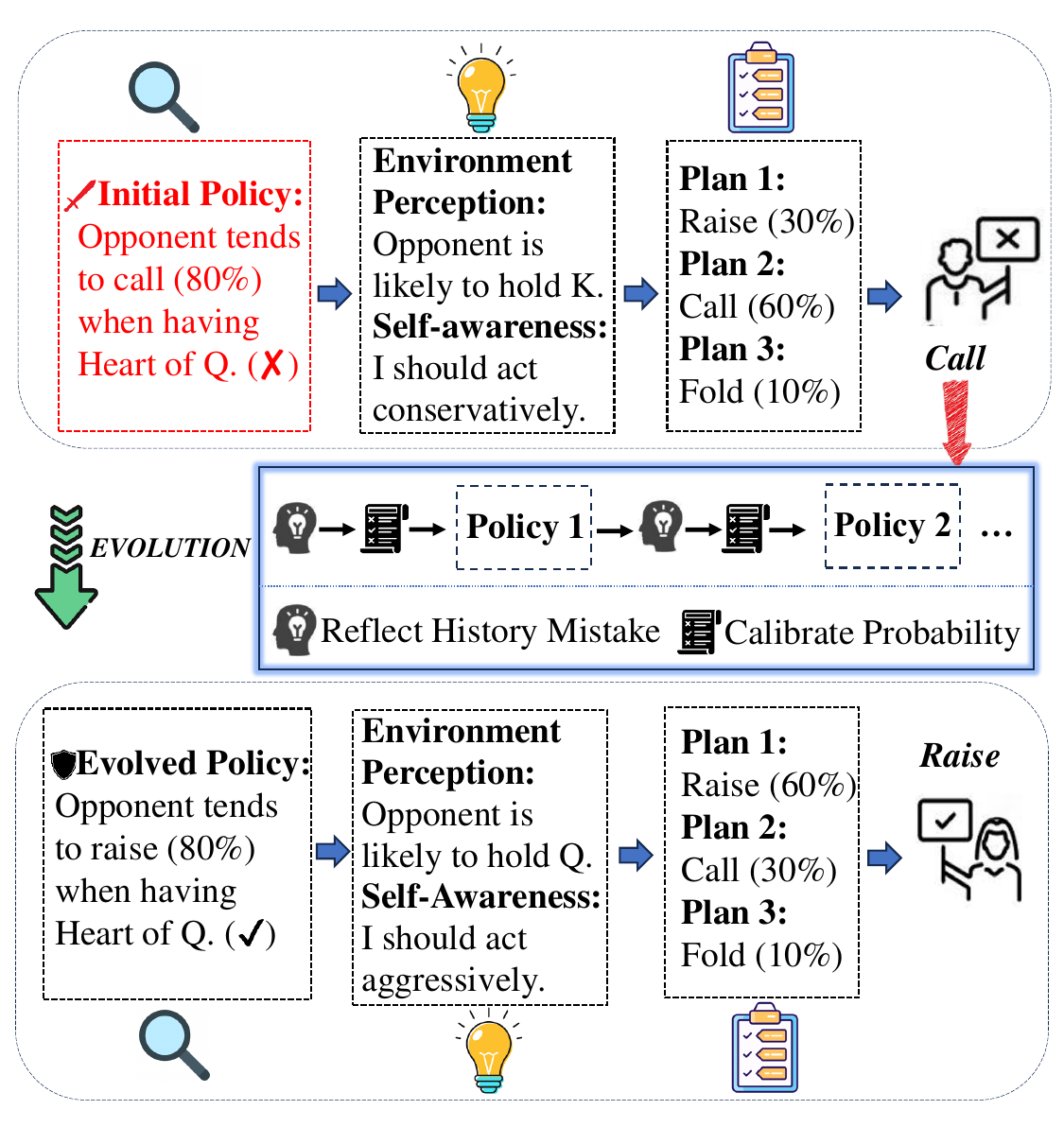}
\vspace{-5pt}
  \caption {Examples of PolicyEvol-Agent’ cognitive process reacting to Opponent's \textit{Raise} Action. The top part shows that the agent made a wrong decision with policy not evolving yet, while the bottom part illustrates the results of reasoning, planning and decision-making reaped from the calibrated policy. We introduce the process of policy evolution in the middle part.}
\vspace{-10pt}
\label{fig1:example}
\end{figure}

Despite markedly superior performance of LLMs, building a robust LLM-based multi-agent to tackle diverse and complicated game simulations remains a challenging endeavor. Firstly, a large variety of game simulations are intricate with dynamic and incomplete information, indicating great efforts to develop cognitive chains \citep{10.1145/3649921.3650013,guo2024largelanguagemodelbased,Gallotta2024LargeLM}. Generally, numerous game simulation tasks involve collaboration and coordination that are beyond the capability of a single agent.
Secondly, most LLM-based agents are incapable of evolving policy comprehensively through post-hoc bias reflections where human could adjust strategies flexibly in case of previous similar decision-making mistakes. In fact, in the dynamic belief generation stage, iterative reflection on past trajectories plays an essential role in psychological judgments of other players \citep{xi2024agentgym,zhang-etal-2024-agent,yuan2024evoagent}.
Thirdly, numerous LLM-based game agents enhance psychological perception via isolated feedback (\emph{e.g.}, environment observation or self-awareness), reducing multifaceted evaluations where human could potentially explore both external and internal beliefs \citep{xi2024agentgym, tao2024surveyselfevolutionlargelanguage,song-etal-2024-trial}.

To address aforementioned points, there has been an increasing trend to develop agent frameworks heavily relying on natural language communication in a sandbox environment within different game settings \citep{10.1609/aiide.v19i1.27534,Lan2023LLMBasedAS,gong-etal-2024-mindagent,mao-etal-2025-alympics}. Various game theory hypotheses, including reasoning, cooperation, persuasion, deception, leadership, etc, are tested or explored in these studies.

We notice that during imperfect information games, human conducts rational maneuvers reaped from a combination of vast memories, strategies, beliefs and plans, all of these constructing their unique characteristics and behavioral patterns. By orchestrating with beliefs, plans, and actions, human can progressively form a global behavioral policy to steer truth-grounded game decisions in the uncertainty scenarios. For the sake of migrating human capabilities to LLMs, Theory of mind (ToM) can be employed to infer environmental uncertainty and understand others' intentions. This cognition pattern provides some insights to propose our game agent for handling complicated missions \citep{Premack_Woodruff_1978,kim-etal-2023-soda,li-etal-2023-theory,Guo2023SuspicionAgentPI,wu-etal-2024-coke}. More detailed works about LLM and ToM can be referred to Appendix~\ref{sec:Related Works}.

In this paper, we introduce \textbf{PolicyEvol-Agent}, a novel method which could elevate behavioral patterns through interacting with the others' stands in confrontation games characterized by solely getting access to individual condition and public observations. Specifically, to align with human maneuvers, we leverage ToM as the chained cognitive process including policy evolution, reasoning, planning, decision-making and reflection. Our LLM-based agent can not only optimize its potentially behavioral policy as the game advances, but also obtain state perception of himself and environment (i.e. other players) in the dynamically interactive context. On the one hand, policy evolution mechanism is endowed with the capacity to calibrate expertise bias based on autonomous reflection on past trajectories without parameter tuning. On the other hand, decisions are ultimately grounded by reacting to those internal and external feedbacks. We instantiates the cognitive process of PolicyEvol-Agent against one opponent in the game as Figure~\ref{fig1:example} shows.

The experiments conducting on Leduc Hold’em \citep{10.5555/3020336.3020403} demonstrate advantages of PolicyEvol-Agent in the game’s winning rates compared to traditional and agent-based models. It also shows the superiority of policy evolution without human guidance throughout the entire game.

The contributions of this work are summarized as follows:
\begin{itemize}
\item We propose PolicyEvol-Agent, a framework encompassing \textbf{memorizing, reasoning, planning, decision-making and reflecting} within dynamic and incomplete information games.

\item \textbf{Policy evolution mechanism} paves the way for adjusting patterns towards a correct direction by integrating game experiences and iteratively updating instructions, encouraging transformation from the novice to an elite.

\item We introduce multifaceted belief generation (i.e. \textbf{self-awareness and environmental perception}), enhancing capabilities for providing rational plans and making feasible decisions.

\item Our proposed PolicyEvol-Agent evaluated on imperfect information games defeats all strong baselines. It exhibits \textbf{strategic human-like maneuvers}, \emph{e.g.}, bluffing, deceiving and flexibly folding.
\end{itemize}
\section{Problem Formulation}
\label{Preliminaries}
\begin{figure*}[t]
  \includegraphics[width=\linewidth]{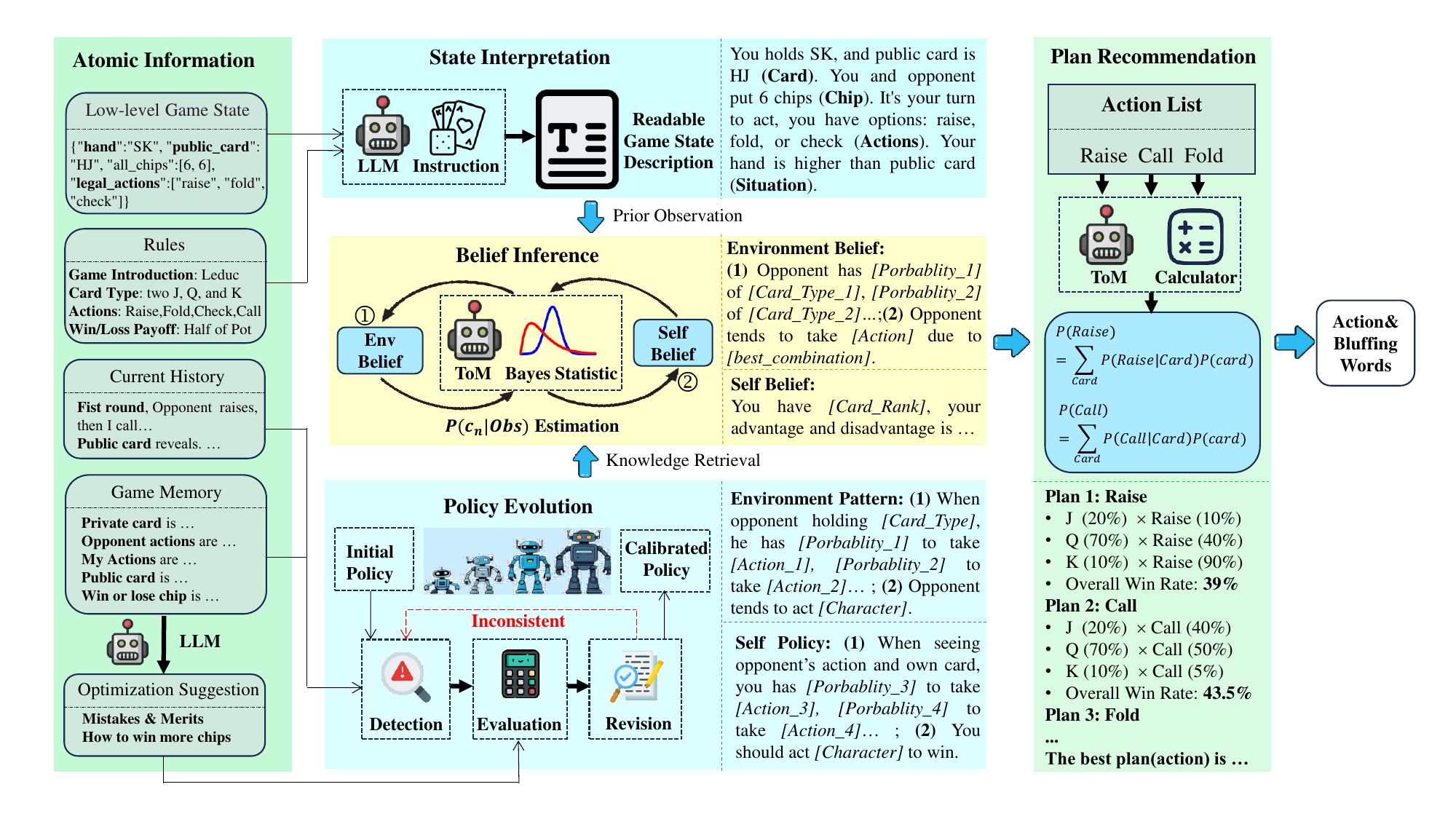}
\vspace{-5pt}
  \caption {Illustration of PolicyEvol-Agent with its four modules detailed. Each module is attached with its cognitive operations and an example output.}
\label{fig2:model}
\vspace{-10pt}
\end{figure*}
 In the dynamically interactive games with incomplete information, players are required to choose rational actions from a valid action list for final victory without knowing opponents' states. Generally, the game process can be formulated as follows from the perspective of one of players. Given the environment set $\mathcal{E}$ and a specific game process $e \in \mathcal{E}$, current observation $o_{i} \in \mathcal{O}$ is either initialized at the beginning of game $(i=1)$ or induced from feedback of previous rounds $(i>1)$ before the player takes action $a_{i} \in \mathcal{A}$ at step $i$, where $\mathcal{O}$ and $\mathcal{A}$ are the observation space and the action space, respectively. Once $a_{i}$ is produced based on multifaceted observation and systematic analysis, other players sequentially react to him, and next new observation $o_{i+1}$ continues. The trajectory is denoted as $t = (o_{1}, a_{1},...,o_{i}, a_{i},...)$ constantly expand through above-mentioned repetition till the game $e$ comes to an end.

In general, LLM-based agents are expected to facilitate the transformation from observation to action by conducting reasoning, planning and decision-making operations in face of uncertainty of opponents' cards. The goal is to find the best trajectory by maximizing the reward function $r(t|e)$:
\begin{equation}
\underset{t}{\operatorname{argmax}}\: r(t|e) = \prod_{i} \pi^{*}(a_{i}|o_{i},h_{i})  \label{reward}
\end{equation}
where action $a_{i}$ is generated through the optimal strategy $a_{i} \sim \pi^{*}(\cdot|o_{i},h_{i})$. $h_{i}$ represents the intermediate cognitive procedures provided by LLMs at step $i$.

In this paper, our study focuses on the Leduc Hold’em simulation, a kind of two-player imperfect information games, and explores the potential of policy evolution for progressively optimizing $\pi^{*}(\cdot|o_{i},h_{i})$ over many games.
\section{Methods}
\label{Methods}
The overview of PolicyEvol-Agent  is illustrated in Figure~\ref{fig2:model}. This innovative agent, without any extra training, comprises four primary modules: Observation Interpretation, Policy Evolution, Multifaceted Belief Generation and Plan Recommendation. To enhance environmental and self perception, we integrate Theory of Mind (ToM) into the latter three modules.
\subsection{Observation Description and Rule Understanding}
State Interpretation aims at transforming game states into a descriptive context with the help of LLM interpretability. The prompt instructions consist of initial state information, observation rules and game rules, converting the current situation into a readable comprehensive text.

Some works \citep{DBLP:journals/corr/abs-1910-04376,Zhao2023DanZeroDT,Guo2023SuspicionAgentPI} report that game state is represented as low-level variables such as a dict in most incomplete information games. In general, the original variables informs both private and public cards, and a valid action list, etc. However, it is hard to understand. Thus observation rules are provided to give a structural explanation to game states. The observation rule instructions can be summarized as follows:

(1) \textbf{Explanation}: To elaborate the variable type of inputs, such as dictionaries, lists or other formats to describe the meaning of elemental names in inputs, for example, "valid$\_$action$\_$list" represents the actions agent can take;
(2) \textbf{Element Descriptions}: To introduce the representation of each element, for example, $\{$"valid$\_$action$\_$list" : ["raise","fold","call"]$\}$ means that agent can choose raise,fold or call;
(3) \textbf{Conversion Tips}: More guidelines for transforming low-level game states into text, for example, SJ denotes the J of Spades and HK denotes King of Hearts.

Game rules are served as means of enhancing LLM understanding for panoramic game introduction, with instructions including general rules, action introduction, payoff principles, and win/loss rules:

(1) \textbf{General Rules}: Brief game introduction (including number of rounds and betting rules): the deck consists only two cards of King, Queen and Jack, totally six cards;
(2) \textbf{Action Descriptions}: Introduction to each actions: players can choose four actions: raise, call, check and fold, and raise action is \emph{[Action Description]};
(3) \textbf{Single Win/Loss Rule}: Requirements for winning, losing, or drawing in a single game: the player whose hand is the same rank as the public card is the winner;
(4) \textbf{Win/Loss Payoff Rule}: Rewards or penalties for winning or losing in a single game: half of the total pot;
(5) \textbf{Whole Win/Loss Rule}: The number of games and the overall win/loss requirements: you aim to have more chips than original after 100 games.

By employing game rules and conversion rules, the low-level game states can be effectively converted into a readable text (denoted as $Obs$). This process is aligned with human preference, and therefore, enhances the comprehensibility of current state. The conditional distribution for each element $Obs[j]$ within the generated text can be modeled as:
\begin{equation}
\begin{split}
Obs\: \sim  \prod^{L}_{j=1} LLM_{\theta}(&Obs[j]|Prompt,Rules,\\
&Obs[0,1,\cdots,j-1])
\end{split}
\end{equation}
where $LLM_{\theta}(\cdot)$ is the large language model parameterized by $\theta$, and $L$ represents the length of the textual elements.

Observation Interpretation facilitates understandable interactions in the gaming environment.  A detailed converted text example can be referred to Appendix~\ref{State Description}.
\subsection{Evolution through Memory and Reflection}
\label{policy evolution section}
In the first few games, it is challenging to discern opponent's behavioral patterns as well as to form one's own game-play tendency. For instance, the agent is required to explore whether opponent bluffs and to react conservatively or aggressively for better decision-making. However, most agents are not capable of adjusting strategies flexibly, lacking understanding of environment and self as game progresses.

Inspired by human-like players, we introduce policy evolution mechanism, characterized by dynamically adjusting strategies by integrating game history and reflection. Assuming that a policy has been obtained in previous games, the Policy Evolution module is designed to calibrate biased probability of policy with the help of instant feedbacks rather than generate it from scratch.

A panoramic policy is the combination of different conditional probability distribution and can be formulated as:
\begin{equation}
Policy \sim \prod^{M,N}_{m=1,n=1} P(a_{m}|c_{n})
\end{equation}
where $a_{m}$ is the $m$th action, $c_{n}$ represents the $n$th card type, $M$ and $N$ denotes the number of legal actions and card types, respectively. The goal of evolution is to fine-tune $P(a_{m}|c_{n})$ by reflecting irrational decisions in game memories until convergence. As Figure~\ref{fig2:model} shows, we summarize the evolution process as three steps: \textbf{Detection}, \textbf{Evaluation} and \textbf{Revision}.

Given the certain old-policy probability $P_{old}(a_{m}|c_{n})$ and updated game histories, the agent firstly calculates the detection probability $P_{detect}(a_{m}|c_{n})$, that is:
\begin{equation}
P_{detect}(a_{m}|c_{n}) = \frac{P(a_{m},c_{n}|History)}{P(c_{n}|History)}
\end{equation}
where ${P(a_{m},c_{n}|History)}$ and ${P(c_{n}|History)}$ can be statistically out.

If $P_{old}(a_{m}|c_{n})$ is inconsistent with what the game memory shows (\emph{i.e.}, $P_{old}\neq P_{detect}$) , then LLM with ToM capability is employed to evaluate the feasible joint of $a_{m}$ and $c_{n}$ based on reflective suggestions as follows:
\begin{equation}
\begin{split}
P(a&_{m},c_{n}|Reflection)= LLM_{ToM}(\\
&P_{old}(a_{m}|c_{n}),History, Reflection)
\end{split}
\label{policy  evolution equation}
\end{equation}
where $P_{old}(a_{m}|c_{n})$ acts as an indicator for irrational deduction, while $Reflection$ includes opponent motivation of each step and self gain or loss. Game reflection points out the excellent or irrational actions, which paves the way for evolutionary direction. Game history can be easily obtained, however, the most essential is to retrospect and gain something empirical, especially when opponent's private card is revealed. The reflective text is generated from:
\begin{equation}
Reflection = LLM_{\theta}(History)
\end{equation}

The detailed instructions of $LLM_{ToM}$ consist of: (1) \textbf{Old Policy}: Including biased self pattern and environmental pattern: opponent is aggressive/conservetive, and agent tends to act when it has different card types;
(2) \textbf{Previous Game Feedback}: Game history summarization, reason for win/lose, and improvement-oriented suggestions: where the agent did wrong or right,  winning/losing more chips if changing to another action.

In fact, the second step of policy evolution involves two parts: environmental pattern evaluation and self pattern evaluation. Environmental pattern stems from experiences, then self pattern additionally takes new environmental pattern into account, which conforms to real-word intuitions. Therefore, Equation~\ref{policy  evolution equation} can be further elaborated and divided into followings:
\begin{equation}
\begin{split}
P&att_{Env} = LLM_{ToM}(Prompt_{patt},\\
& Rules, History, Reflection, P_{old}, Obs)
\end{split}
\end{equation}
\begin{equation}
\begin{split}
P&att_{Self}= LLM_{ToM}(Prompt_{patt}, Patt_{Env} \\
& Rules, History, Reflection, P_{old}, Obs)
\end{split}
\end{equation}
where $Prompt_{patt}$ is the pattern prompt, $P_{old}$ represents previous policy. $LLM_{ToM}(\cdot)$ denotes the large language model with ToM inference similar to Suspicion-Agent \citep{Guo2023SuspicionAgentPI} and Agent-Pro \citep{zhang-etal-2024-agent}. $Patt_{Env}$ pictures opponent actions in certain card type, and derives its overall character (aggressive or conservative). $Patt_{self}$, on the other hand, is a response to $Patt_{Env}$, including bluffing or not, and action tendencies with different conditions.

After joint evaluation of actions and card types (see Equation~\ref{policy  evolution equation}), the final revised policy is calculated by:
\begin{equation}
P_{new}(a_{m}|c_{n}) \approx \frac{P(a_{m},c_{n}|Reflection)}{P(c_{n}|History)}
\end{equation}

$P_{new}$ not only provides guidelines for belief inference (see Section~\ref{Belief Generation Section}), but also serves as the inheritable framework for next game policy modification with only replacing irrational deduction. 
The prompt template for policy evolution can be found in Appendix~\ref{policy evolution template}.
\subsection{Environmental and Self-aware Belief Generation}
\label{Belief Generation Section}
Before making decisions in a long-term interactive process, it is necessary to ensure belief rationality. In view of ToM reasoning in imperfect information games, PolicyEvol-Agent is instructed to explore potential information based on the prior observation and the revised policy. In other words, this process aims to precisely estimate the conditional probability $P(c_n|Obs)$ by
\begin{equation}
P(c_n|Obs)\approx\frac{P(c_n, a_m|History, Obs)}{P(a_m|c_n, Policy)}
\end{equation}
where $P(a_m|c_n, Policy)$ represents knowledge retrieval from the calibrated policy directly.

Similar to Policy Evolution module, the belief perception consists of self-awareness and environmental reasoning. With the help of $LLM_{ToM}(\cdot)$, two parts $Belief_{Env}$ and $Belief_{Self}$can be formulated as:
\begin{equation}
\begin{split}
Bel&ief_{Env} = LLM_{ToM}(Prompt_{belief},\\
&Rules, Reflection, Patt_{Env},Obs)
\end{split}
\end{equation}
\begin{equation}
\begin{split}
Bel&ief_{Self}= LLM_{ToM}(Prompt_{belief},Rules,\\
&Reflection, Patt_{Self}, Belief_{Env}, Obs)
\end{split}
\end{equation}
where $Prompt_{belief}$ denotes the prompt template for LLM-based belief inference. The participation of $Patt_{Env}$ and $Patt_{Self}$ provides insights for retrieval-augmented belief. Similar to Section~\ref{policy evolution section}, $Belief_{Env}$ is the foundation of $Belief_{Self}$.
The key information of $Belief_{Env}$ and $Belief_{Self}$ summarizes the following contents:

(1) \textbf{Environmental Belief}: Card type with probability, and the best possible combination of opponent;
(2) \textbf{Self-aware Belief}: Advantages and disadvantages with self probability for current goal and long-term goal.

We also provide a belief example in Belief Inference module of Figure~\ref{fig2:model}. It is obvious that PolicyEvol-Agent will provide more reasonable and reliable plans as a result of these grounded beliefs (see Section~\ref{plan section}). The Belief Generation prompt template can be seen in Appendix~\ref{Belief Generation prompt}.
\subsection{Plan Recommendation with Probability}
\label{plan section}
Under guidance of the comprehensive observations and the rational beliefs, LLM is capable of recommending plans attached to winning rate. Based on a legal action list, LLM is employed to evaluate each action by calculating expectation of potential chip gains, and then summarize these probabilities into specific instructions. To enhance the controllability of the interactive agent, we also introduce a critic instruction to select proposals and to provide feedbacks to the role-paying agent. We prompt LLM to make text-based plans as:
\begin{equation}
\begin{split}
[Pl&ans, Plan_{Best}]= LLM_{ToM}(Obs,\\
& [Belief_{Env},Belief_{Self}], Reflection)
\end{split}
\end{equation}

The recommended plans are exhibited as follows:

(1) \textbf{Plan 1}: when I raise, opponent card type is \emph{[Card Type]}, the average of win rate is \emph{[Probability]};
(2) \textbf{Plan 2}: when I call, opponent card type is \emph{[Card Type]}, the average of win rate is \emph{[Probability]};
(3) \textbf{Plan 3}: when I fold, opponent card type is \emph{[Card Type]}, the average of win rate is \emph{[Probability]};
\textbf{Best Plan}: Choose the bast plan and explain why: plan 2 (call) is best because it has the highest rate from a/an aggressive/conservative perspective.

Then the action can be triggered by:
\begin{equation}
a_{i}=LLM_{\theta}(Obs,[Plans, Plan_{Best}])
\end{equation}
where $a_{i}$ represents the $i$-th step action.
\begin{table*}[t]
\centering
\begin{tabular}{lcccccccc}
\hline
    \multirow{2}{*}{Agent Model} & \multicolumn{6}{c}{Opponent Model} & \multirow{2}{*}{Avg.} & \multirow{2}{*}{Win Rate} \\
\cmidrule{2 - 7}
    & NFSP & DQN & DMC & CFR & Susp & PolicyEvol &  &  \\
\hline
NFSP & - & -33 & -22 & -45 & -96 & -123 & -63.8 & 0\% \\
DQN & +33 & - & -55 & -20 & -7 & -21 & +14 & 20\% \\
DMC & +22 & +55 & - & +16 & -22 & -12 & +11.8 & 60\% \\
CFR & +45 & +20 & -16 & - & +15 & -35 & +5.8 & 60\% \\
Susp (GPT 3.5) & +3 & -200 & +49 & -73 & - & - & -55.25 & 50\% \\
Susp (GPT 4) & +142 & +45 & +24 & +37 & - & - & +62 & 100\% \\
\hline
Susp (qwen) & +96 & +7 & \textbf{+22} & -15 & - & -41 & +13.8 & 60\% \\
\textbf{PolicyEvol (qwen)} & \textbf{+123} & \textbf{+21} & +12 & \textbf{+35} & \textbf{+41} & - & \textbf{+46.4} & \textbf{100\%} \\
\hline
\end{tabular}
\vspace{-5pt}
\caption{The performance of PolicyEvol-Agent when fighting against different methods after 100 games on Leduc Hold’em environment. The results report the total win/lose chips after playing of each method, and the number of win/lose chips in each game ranges from 1 to 14. "Susp" denotes Suspicion-Agent.}
\vspace{-10pt}
\label{main result table}
\end{table*}
\section{Experinments}
\subsection{Task and Dataset}
To demonstrate the performance of PolicyEvol-Agent in incomplete information environment characterized by only having access to self and public information without others' state, we choose Leduc Hold’em as the experimental game with RLCard framework \citep{DBLP:journals/corr/abs-1910-04376}. As a simplified version of Limit Texas Hold’em,  there are four optional actions in Leduc Hold’em: \texttt{Raise}, \texttt{Call}, \texttt{Fold} and \texttt{Check}. The number of information sets is about $10^2$, and the average number of states in a single information set is $10^2$. The detailed game introduction including game rules and win/loss rules can be seen in Appendix~\ref{sec:Detailed Game Information}.

We conduct simulation experiments based on the open-source language model \texttt{qwen-max} \citep{Bai2023QwenTR,yang2024qwen2} to compete against other agent opponents. Following \citet{10.5555/3020336.3020403,Guo2023SuspicionAgentPI}, we select second-order ToM for reasoning, planning and decision-making manipulations. At the end of each game, opponent’s card is revealed for better game history reflection.
\subsection{Baseline}
\begin{table}
\centering
\begin{tabular}{lccc}
    \hline
  \multirow{2}{*}{Strategy} & \multicolumn{3}{c}{Opponent Model} \\
       \cmidrule{2 - 4}
  & CFR & DMC & Suspicion\\
        \hline
    Ours & +35 & +12 & +41\\
    \:--w/o Policy & +6 & -23 & -28\\
    \:--w/o Belief & -13 & -30 & -21\\
     \:--w/o Plan& -21 & -56 &-44\\
     \:--w/o Reflection & -7 & -16 & +19\\
    \hline
\end{tabular}
\vspace{-5pt}
\caption{Comparison of PolicyEvol-Agent without Policy, Belief, Plan or Refelction when competing with CFR, DMC and Suspicion-Agent on Leduc Hold’em environment. The total win/lose chips against these methods are reported after 100 games.}
\vspace{-10pt}
\label{ablation study table}
\end{table}
We select a wide range of traditional methods and agent-based models as competitive opponents. All these role-playing methods exhibit the adaptability at a human level. \textbf{Suspicion-Agent} \citep{Guo2023SuspicionAgentPI} is a state-of-the-art model and serves as the upper bound compared with other approaches. To our knowledge, Suspicion-Agent is the first text-oriented agent for Leduc Hold’em without specific training, which integrates ToM reasoning, planning and decision-making capabilities to influence players' behavior. Other traditional algorithms includes \textbf{NFSP} \citep{Heinrich2016DeepRL}, \textbf{DQN} \citep{Mnih2015HumanlevelCT}, \textbf{Deep Monte Carlo Search (DMC)} \citep{pmlr-v139-zha21a} and \textbf{CFR} \citep{Zinkevich2007RegretMI}. NFSP and DMC are based on self-play, specifically developed for incomplete information games. DQN allows a reinforcement-learning agent to learn policies directly from high-dimensional sensory input. And CFR is a method with game theory. Each of them has potential decision-making capabilities in imperfect information games.
\subsection{Evaluation Metrics}
To evaluate the performance of PolicyEvol-Agent, 100 new betting game hands are randomly sampled and allocated to two sides during the game battle, and then the total chips remained are counted for final victory. The policy are continuously fine-tuned along with the 100 games. We investigate a variety of works related to evaluation metrics for game agents in Appendix~\ref{sec:Related Works}, and the evaluation method in this paper follows \citet{Guo2023SuspicionAgentPI}.
\subsection{Main Results}
As shown in Table~\ref{main result table}, we report the final chips of our LLM-based PolicyEvol-Agent against the other players (NFSP, DQN, DMC, CFR and Suspicion-Agent). For the sake of better comparison, we also reproduce the results of Suspicion-Agent competing with other methods with the help of texttt{qwen-max}. The results indicate the superiority of our model in game simulation scenarios.

PolicyEvol-Agent gains remarkable chips compared with numerous approaches including RL-based algorithms and the other state-of-the-art LLM-based agent. We observe that PolicyEvol-Agent with \texttt{qwen-max} surpasses all the listed methods in Table~\ref{main result table}. In particular, it beats Suspicion-Agent with \texttt{qwen-max} by a large margin (41 chips) after 100 random game simulations. Besides, our designed agent outperforms other traditional methods by winning a large amount of chips (123, 21, 12 and 35 chips respectively). These findings demonstrates the strengths of policy evolution and ToM capability of LLM in incomplete information games, as well as the effectiveness of environmental perception and self-awareness in dynamic interactions.

PolicyEvol-Agent has a better performance than Suspicion-Agent when employing the same large language models. It is evident that PolicyEvol-Agent achieved more chips than Suspicion-Agent with \texttt{qwen-max} when competing with NFSP, DQN and CFR, though the winning chips of PolicyEvol-Agent against DMC are slightly lower than that of Suspicion-Agent (-10 chips). In addition, the performance of PolicyEvol-Agent with \texttt{qwen-max} lies between Suspicion-Agent with GPT 3.5 and GPT 4. This phenomenon may be caused by the differences among versions of large language models.
\begin{figure*}[t]
  \includegraphics[width=0.47\linewidth]{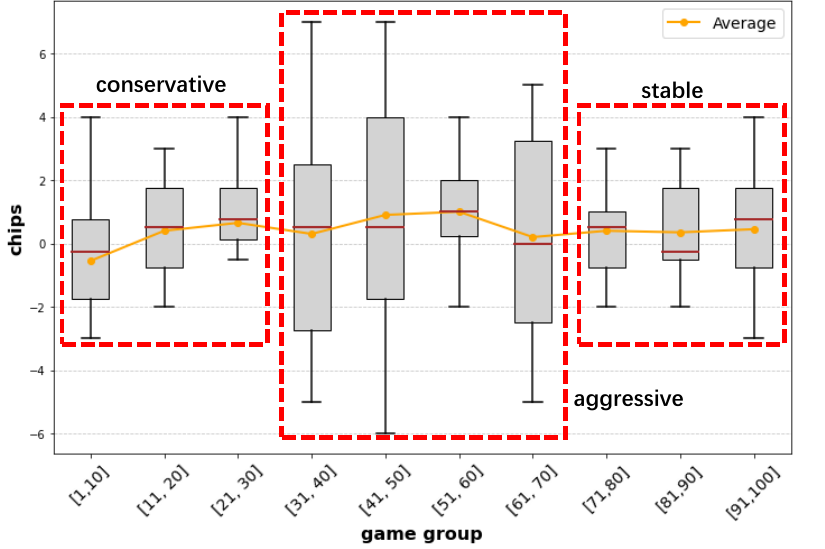}\hfill
  \includegraphics[width=0.48\linewidth]{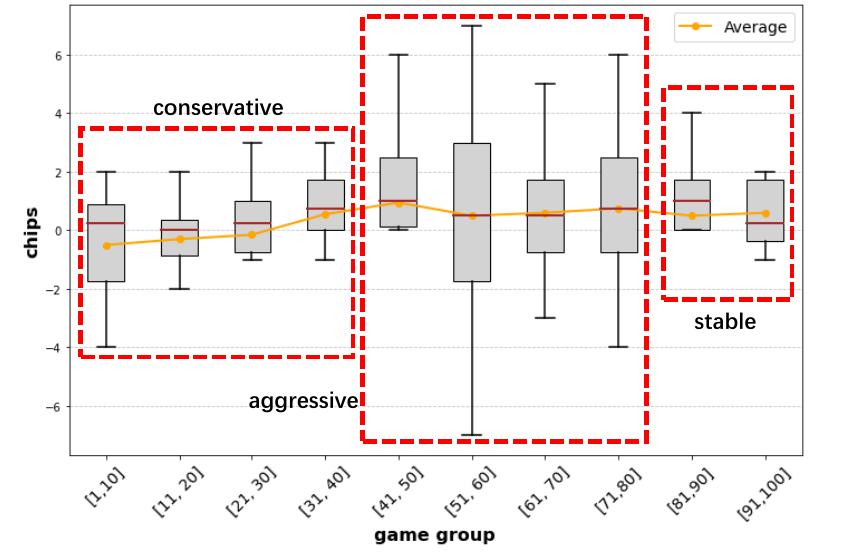}
  \vspace{-5pt}
\caption {Chip gains of each ten games during the evolution process. \textbf{Left figure: }PolicyEvol-Agent \emph{vs.} Suspicion-Agent. \textbf{Right figure: }PolicyEvol-Agent \emph{vs.} CFR. We illustrate the average and median chip gains in orange line and gray box respectively.}
\vspace{-10pt}
\label{evolution figure}
\end{figure*}
\subsection{Ablation Study}
We further investigate the effectiveness of four essential components on PolicyEvol-Agent performance: Policy Evolution, Belief Generation, Plan Recommendation and Reflection module. The chip gain results against CFR, DMC and Suspicion-Agent over 100 games are presented in Table~\ref{ablation study table}.

Planning Recommendation has the biggest influence on boosting the final performance. Without recommendation, PolicyEvol-Agent can be confused and eventually select an action from the action list randomly, because planning Recommendation points out the best action to the LLM-based agent straightforwardly.

With the help of Belief Generation, PolicyEvol-Agent emerges remarkable advantages. The capabilities of this LLM-based agent can be weakened more or less because it is challenging to mining potential information in game simulations.

Policy Evolution also plays a vital role in winning more chips in incomplte information scenarios. Even though the game environment provides LLMs with a list of valid actions, the agent cannot effectively make a rational decision due to lacking reasoning on previous reflection.

Reflection behaves the least in game evaluations. We find that evolutionary policy and multifaceted beliefs may cover a proportion of reflective information, which weakens the influence of Reflection module. That is why Reflection has a smaller advantage than other components.
\subsection{Analysis on Evolution Process}
To visualize the policy evolution process, we depict the chip gain results in every 10 games in Figure~\ref{evolution figure}. During the 100 games, the entire evolution process can be roughly divided into three stages when competing with Suspicion-Agent and CFR: \textbf{conservation, aggression and stability}.

The conservation stage aims to observe opponent's reaction pattern and lose chips as less as possible. However, due to lack of comprehensive perception towards opponent, it is \textbf{inevitable to make irrational decisions and lose chips}.
\begin{figure}
  \includegraphics[width=\columnwidth]{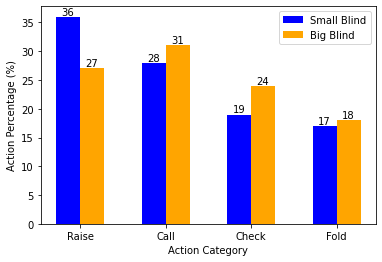}
  \caption {Proportion of different actions taken by PolicyEvol-Agent in small blind and big blind in 50 games when fighting against Suspicion-Agent.}
  \vspace{-10pt}
  \label{blind}
\end{figure}

After a certain amount of games, PolicyEvol-Agent can gain some historical experiences and initially form an effective policy to defeat opponents. As a result of drastic policy updating (40-70th games in general), PolicyEvol-Agent behaves aggressively and occasionally makes wrong decisions in this stage. Thus, the aggression stage is characterized by \textbf{high rewards and instability}.

The third stage shows that PolicyEvol-Agent has a good command of the psychological state of opponent and its own. In this stage, PolicyEvol-Agent is capable of reasoning intentions of opponent and contrlling its own chip losses. In addition, we exhibit some game histories in the Appendix~\ref{sec:Case Study}. By referring to these cases, we can find that the policy has been \textbf{refined comprehensively} in the stable stage.
\subsection{Small Blind \emph{vs.} Big Blind}
Small blind means that the player acts first, while big blind puts more chips than small blind and acts next. In order to explore the action preference of PolicyEvol-Agent in different blind position, we calculate the proportion of different actions taken by PolicyEvol-Agent at different positions in 50 games (see Figure~\ref{blind}).

It can be observed that PolicyEvol-Agent tends to act \textbf{more aggressively when positioned at small blind}. When PolicyEvol-Agent acts first, it takes more \texttt{raise} actions than \texttt{check} and \texttt{call} actions. When it comes to act at big blinds, the situation is on the contrary. This indicates that PolicyEvol-Agent is equipped with a bluffing strategy to stand at a dominant position, so as to win more and loss less even under adverse conditions.
\section{Conclusion}
In this paper, we develop a LLM-based agent, PolicyEvol-Agent, equipped with psychological state perception and capable of evolving biased policy in incomplete information games with Theory of Mind. It dynamically interacts with environment and iteratively calibrates previous policy by learning from historical experiences. The environmental and self-aware beliefs are generated based on the optimized policy and ToM capablity of LLM. We evaluate PolicyEvol-Agent through competing with a series of traditional methods and the strong baseline Suspicion-Agent in the two-player game Leduc Hold’em. The significant chip gains show the comprehensive capabilities of PolicyEvol-Agent in reasoning, planning, decision-making and planning.
\section*{Limitations}
PolicyEvol-Agent has presented a novel cognition chain in dynamically interactive games, but we want to declare that there remain limitations and improvement.

\textbf{Diversify Game Scenarios.}  PolicyEvol-Agent only has access to two-player confrontational games. However, there are a wide range of complex games in the real word. Expanding its scope to multi-player cooperative and competitive games would enhance its practicality. For example, in team-based strategy games, agents need to communicate, coordinate, and make decisions. Additionally, exploring real-time game settings would further challenge and refine its cognitive abilities, as different pacing and decision-making timelines would come into play.

\textbf{Enhance the applicability of LLMs.} Due to financial budget, we select suitable LLMs from an economical perspective. However, different versions of LLMs have different levels of understanding of game scenario. Therefore, it is necessary to analyze the performance of various LLMs first, not only considering their accuracy in interpreting game rules and predicting opponent behaviors but also factoring in their computational efficiency and resource consumption. Especially in games with rich narrative contexts, the LLMs that excel in natural language generation to facilitate more diverse interactions are prioritized.
\section*{Ethics Statement}
Our research was conducted with the objective of boosting the applicability of multi-agent systems. We must declare that our work adheres strictly to legal and ethical standards. Throughout the study, all experiments are performed in safe game environments, ensuring that there will be no negative impact on society. We recognize the potential risks associated with teenager game addiction and improper profit-making purpose. Therefore, we have implemented strict principles to mitigate any unintended consequences. We are devoted to develop our technologies to contribute positively to the multi-agent systems, and we advocate for the ethical use of gaming AI agents.
\bibliography{arxiv_main}
\appendix
\section{Case Study}
\label{sec:Case Study}
In this section, we provide several game history respectively between PolicyEvol-Agent and Suspicion-Agent to show the process of policy evolution through behavioral patterns. In these histories, "board\_game\_expert" is PolicyEvol-Agent, and "GoodGuy" is Suspicion-Agent.
\begin{lstlisting}
7th Game Starts.
GoodGuy has the observation {'hand': 'HK', 'public_card': None, 'all_chips': [2, 1], 'my_chips': 1, 'legal_actions': ['call', 'raise', 'fold']}, tries to take action: call.
board_game_expert has the observation: {'hand': 'HJ', 'public_card': None, 'all_chips': [2, 2], 'my_chips': 2, 'legal_actions': ['raise', 'fold', 'check']}, tries to take action: check, and says "I'll check and see what the public card brings." to GoodGuy.
GoodGuy has the observation: {'hand': 'HK', 'public_card': None, 'all_chips': [2, 1], 'my_chips': 1, 'legal_actions': ['call', 'raise', 'fold']}, and tries to take action: call.
board_game_expert has the observation: {'hand': 'HJ', 'public_card': 'HQ', 'all_chips': [2, 2], 'my_chips': 2, 'legal_actions': ['raise', 'fold', 'check']}, tries to take action: check, and says "I'll check and see what you do next." to GoodGuy.
GoodGuy has the observation: {'hand': 'HK', 'public_card': 'HQ', 'all_chips': [2, 2], 'my_chips': 2, 'legal_actions': ['raise', 'fold', 'check']}, says "I think I'll just see what you do next." to board_game_expert, and tries to take action: check.
GoodGuy win 1.0 chips, board_game_expert lose 1.0 chips.
\end{lstlisting}
\begin{lstlisting}
30th Game Starts.
GoodGuy has the observation {'hand': 'SK', 'public_card': None, 'all_chips': [2, 1], 'my_chips': 1, 'legal_actions': ['call', 'raise', 'fold']}, tries to take action: raise
board_game_expert has the observation: {'hand': 'HK', 'public_card': None, 'all_chips': [2, 4], 'my_chips': 2, 'legal_actions': ['call', 'raise', 'fold']}, tries to take action: raise, and says "I think my hand is too strong to pass up this opportunity. Let's see what you've got." to GoodGuy.
GoodGuy has the observation: {'hand': 'SK', 'public_card': None, 'all_chips': [2, 1], 'my_chips': 1, 'legal_actions': ['call', 'raise', 'fold']}, and tries to take action: raise.
board_game_expert has the observation: {'hand': 'HK', 'public_card': 'SJ', 'all_chips': [6, 6], 'my_chips': 6, 'legal_actions': ['raise', 'fold', 'check']}, tries to take action: raise, and says "I think my hand is too strong to pass up this opportunity. Let's see if you're willing to match this." to GoodGuy.
GoodGuy has the observation: {'hand': 'SK', 'public_card': 'SJ', 'all_chips': [10, 6], 'my_chips': 6, 'legal_actions': ['call', 'raise', 'fold']}, and tries to take action: call.
GoodGuy win 0.0 chips, board_game_expert lose 0.0 chips.
\end{lstlisting}
\begin{lstlisting}
70th Game Starts.
board_game_expert has the observation: {'hand': 'HK', 'public_card': None, 'all_chips': [1, 2], 'my_chips': 1, 'legal_actions': ['call', 'raise', 'fold']}, tries to take action: raise, and says "GoodGuy, I'm feeling confident with my hand. Let's see how you respond to this!" to GoodGuy.
GoodGuy has the observation {'hand': 'SJ', 'public_card': None, 'all_chips': [4, 2], 'my_chips': 2, 'legal_actions': ['call', 'raise', 'fold']}, tries to take action: call.
board_game_expert has the observation: {'hand': 'HK', 'public_card': 'SK', 'all_chips': [4, 4], 'my_chips': 4, 'legal_actions': ['raise', 'fold', 'check']}, tries to take action: raise, and says "With two Kings on the board, I think my hand is pretty strong. Let's see how you respond." to GoodGuy.
board_game_expert win 2.0 chips, GoodGuy lose 2.0 chips.
\end{lstlisting}
\begin{lstlisting}
86th Game Starts.
board_game_expert has the observation: {'hand': 'HK', 'public_card': None, 'all_chips': [1, 2], 'my_chips': 1, 'legal_actions': ['call', 'raise', 'fold']}, tries to take action: raise, and says "I think I'll raise here. Let's see what you've got." to GoodGuy.
GoodGuy has the observation {'hand': 'HJ', 'public_card': None, 'all_chips': [4, 2], 'my_chips': 2, 'legal_actions': ['call', 'raise', 'fold']}, tries to take action: call.
board_game_expert has the observation: {'hand': 'HK', 'public_card': 'HQ', 'all_chips': [4, 4], 'my_chips': 4, 'legal_actions': ['raise', 'fold', 'check']}, tries to take action: check, and says "I'll check. Let's see what you've got." to GoodGuy.
board_game_expert win 2.0 chips, GoodGuy lose 2.0 chips.
\end{lstlisting}
We can observe that early PolicyEvol-Agent is less rational, with little reflection at first. As evolution progresses, PolicyEvol-Agent exhibits more aggressively with more \texttt{raise} and less \texttt{fold}. At final stage, it equips flexible and approachable strategy. By comparing the intermediate processes of the these games, it is evident that the evolution of behavior: from chaotic to stable, from irrational to rational.
\section{Detailed Game Information}
\label{sec:Detailed Game Information}
\subsection{Game Introduction}
Leduc Hold'em is first introduced in \citet{10.5555/3020336.3020403} and sometimes used in academic research. It is played with a deck consisting only two cards of King, Queen and Jack, six cards in total. Each game is fixed with two players, only two rounds, and two-bet maximum in each round. The game begins with each player being dealt one card privately, followed by a betting round. Then, another card is dealt faceup as a community (or board) card, and there is another betting round. Finally, the players reveal their private cards. If one player's private card is the same rank as the board card, he or she wins the game; otherwise, the player whose private card has the higher rank wins.
\subsection{Rules}
One public hand is available at the beginning of the second round. There are four actions that players can choose: raise, call, check and fold. Raise action: In the first round (when the public card not revealed), you have to put chips to make sure your chips in pot with amounts of 2 more than that of your opponent when you raise; in the second round (after the public card revealed), you have to put chips to make sure your chips in pot with amounts of 4 more than that of your opponent when you raise. Call action: you will put chips with same amounts as your opponent (if his chips in the pot are higher than yours).  Check action: you will pass on the opportunity to bet and wait to see what the next player does. Fold action: you will drop out and lose the money you may have already put in the pot when you decide not to continue playing the game.  In the first round, one player is randomly choosed to put 1 unit in the pot as small blind while the other puts 2 unit as big blind, and each player is dealt one private card, then starts betting. The player with small blind acts first. In the second round, one public card is revealed first, then the players bet again. (Only one public card is available in the whole game.)

Single Game Win/Draw/Lose Rule: the player whose hand has the same rank as the public card is the winner of one round game. If neither, then the one with higher rank wins this round game, if the rank of cards of two players are the same, it is draw. You can also 'fold' in one round game then your opponent wins the game directly.

Whole Game Win/Draw/Lose Rule: you are requested to attend 100 games with your opponent in the test stage (i.e. ), you both are given 100 chips originally, and the guy who have more chips will win the game after the tested 100 games (i.e. You aim to have more chips than your original chips).

Winning Payoff Rule:  The half of  the total pot.

Lose Payoff Rule:  The half of the total pot.
\section{Related Works}
\label{sec:Related Works}
\subsection{LLM Capabilities}
In recent years, great efforts have been put to exploit the reasoning and planning capabilities of LLMs in a wide range of natural language tasks. To achieve predefined goals, a series of paradigms such as Chain of Thought (CoT), Tree of Thought (ToT) and Graph of Thought(GoT) are proposed to advance prompting capabilities of LLMs by imitating the formulating procedure of human thoughts \citep{wei2022chain, yao2024tree, besta2024graph}. Subsequently, OpenAI introduced the o1 model by embedding the CoT process and integrating reinforcement learning into training to further enhance LLM reasoning performance \citep{jaech2024openai}. Moreover, retrieval-augmented generation (RAG) has showed effectiveness in factual decision making and LLM-generated responses by grounding external resources \citep{lee-etal-2024-planrag,asai2024selfrag}.
To ensure LLMs is capable to perform complex tasks based on reasoning and planning over mere language generation, such as perceiving environment and taking actions, the leverage of collective intelligence and skills of agents is more prevailing. ReACT \citep{yao2023react} synergized both reasoning and acting in question answering and fact verification scenarios. Self-Reflexion \citep{shinn2024reflexion} executed sequential action choices based on feedback from long trajectories and consequentially achieved improvements in decision-making, coding and reasoning tasks. LLM-based agent is also well-applied in simulated gaming scenarios as players. \citet{akata2023playing} leveraged LLMs to play repeated games Battle of the Sexes relying on natural language communication and offering human-like plans. \citet{yim2024evaluating} developed a sophisticated text-based agent by integrating reinforcement learning as an external tool simultaneously in Guandan Card games.
\subsection{Theory of Mind Application}
Theory of Mind (ToM), well-known for its cognitive ability that allows LLMs to perceive and infer environmental and their own states like beliefs, knowledge, desires, and intentions \citep{premack1978does, frith2005theory,hurley2008shared,DBLP:conf/emnlp/ChanJYDF0L0WS24}, has been widely applied to anticipate individuals’ motivation and improve reasoning and decision-making in imperfect information games \citep{de2014theory}. Although ToM of large language models still face challenges in the existing literature,  it has shown great potential in complex text-based real-word simulation including imperfect information game-playing like poker or chess in recent studies \citep{yuan2020emergence,kosinski2023theory,moghaddam2023boosting,zhang-etal-2024-agent}. Suspiicion-Agent \citep{Guo2023SuspicionAgentPI} has a state-of-the-art performance in the Leduc Hold’em, which demonstrates the first-order and second-order ToM inferences involving dynamic interactions. \citet{li-etal-2023-theory} explored LLM-based agents in a multi-agent cooperative text game through explicit belief state representations, finding that it enhances  the accuracy of ToM inferences. \citet{yim2024evaluating} proposed a ToM planning technique that enables embodied agents to collaborate strategically with adversarial agents in Guandan. Inspired
 by these studies, our research aims to improve LLM-based agents’ behavioral patterns by incorporating ToM capabilities into policy evolution.
\onecolumn
\section{Prompts for PolicyEvol-Agent}
\label{sec:Prompt}

\subsection{State Interpretation}
\label{State Description}
\begin{lstlisting}
"You are the player behind a NPC character called {agent_name} attached to player index {user_index}, and you are playing the board game {game_name} against {recipient_name}. \n"
" The game rule is: {rule} \n"
" Your observation now is: {observation}\n"
" And the observation conversion rules are: {observation_rule}\n"
" You will receive a valid action list you can perform in this turn. \n"
" Your valid action list is: {valid_action_list}\n"
" Please convert {observation} and {valid_action_list} to the readable text based on the observation conversion rules and your knowledge about the {game_name} (respond shortly).\n\n"
\end{lstlisting}
\subsection{Policy Evolution}
\label{policy evolution template}
Environmental Pattern Prompt:
\begin{lstlisting}
"You are the player behind a NPC character called {agent_name}, and you are playing the board game {game_name} with {recipient_name}. \n"
 " The game rule is: {rule} \n"
 " Your previous round history in this current game is: {cshort_summarization}\n"
" Your previous game memories including observations, actions, conversations with {recipient_name} and your reflection are: \n{long_memory}\n"
 " Your previous behavioral pattern analysis about your opponent {recipient_name} is: {old_opponent_pattern}\n"
" Please understand the game rule, all previous game summarization and previous game pattern of {recipient_name}, can you do following things for future games? \n"
" Revise {recipient_name}'s game pattern with public card not released: please infer or update all possible reasonable {recipient_name}'s game pattern/preferences for each card he holds with probability (normalize to number 100\% in total for each pattern) as a tree-structure output step by step when public card can't be observed."
 " Output: In the rounds with public card not released, when name holds card1, he would like to do action (probabilities); when name holds card2, he would like to do action (probabilities), ... \n "
 " Judge {recipient_name}'s character: please infer {recipient_name}'s behavioral character based on the newly-updated game pattern and game history. Output: To my konwledge, {recipient_name} tends to act radically/conservatively/neutrally/flexibly."
 \end{lstlisting}
Self Pattern Prompt:
\begin{lstlisting}
 "You are the objective player behind a NPC character called {agent_name}, and you are playing {game_name} against {recipient_name}. \n"
 " The game rule is: {rule} \n"
 " Your previous round history in this current game is: {cshort_summarization}\n"
 " Your previous game memories including observations, actions, conversations with {recipient_name} and your reflection are: \n{long_memory}\n"
 " Your previous behavioral strategy pattern is: {old_self_pattern}\n"
 " Your current behavioral pattern analysis and reasoning about your opponent {recipient_name} is: {opponent_pattern}\n"
 " Please understand the game rule, all previous game records, current game pattern of {recipient_name} and your previous behavioral strategy, can you do following things for future games?\n "
 " Reflection: Reflex which your actions are right or wrong in previous games (especially when the public card does not release), and think why you win or lose concrete chips step by step.  (Note that you cannot observe the cards of the opponent during the game, but you can observe his actions.)\n "
 " Strategy Improvement: Understanding the above information combined with your reflection, think about what strategies I can adopt to exploit the game pattern of {recipient_name}. By taking {recipient_name}'s guess on my game pattern without public card observed into account, revise previous strategies for winning {recipient_name} step by step."
 " Output: When I hold card without public card, and see the action of the opponent, I would like to do action 1 (probabilities), action2 (probabilities) (normalize to number 100% in total), to my konwledge, I tend to act radically/conservatively/neutrally/flexibly, because I can infer that {recipient_name} is (or not) possibly bluffing; continue ... "
 \end{lstlisting}

 \subsection{Belief Generation}
\label{Belief Generation prompt}
Environmental Belief Prompt:
\begin{lstlisting}
"You are the player behind a NPC character called {agent_name}, and you are playing the board game {game_name} against {recipient_name}. \n"
" The game rule is: {rule} \n"
" Your estimated judgement about the behaviour pattern and character of {recipient_name} is: {pattern} \n"
" Your observation now is: {observation}\n"
" Your current game progress summarization including actions and conversations with {recipient_name} is: {recent_observations}\n"
" Your previous game memory including observations, actions, conversations with {recipient_name} and your reflection is: \n{long_memory}\n"
" Understanding the game rule, the cards you have, your observation, progress summarization in the current game and previous game history, the estimated behaviour pattern and character of {recipient_name}, the potential guess pattern of {recipient_name} on you, and your knowledge about the {game_name}, can you do following things?\n "
" Belief on {recipient_name}'s cards: Understanding all given information, please infer the probabilities about the cards of {recipient_name}  (normalize to number 100% in total) objectively step by step."
" Output: {recipient_name} saw my history actions (or not) and then did action1 (probability) in the 1st round , ... continue..... Before this round, {recipient_name}  see my history actions (or not) and  did action1 (probability), because of {recipient_name}'s behaviour pattern and the match with the public card (if release), {recipient_name} tends to have card1 (probability), card2 (probability) ..continue.. (normalize to number 100% in total).\n"
" Analysis on {recipient_name}'s Cards: please analysis what is {recipient_name}'s best combination and advantages of {recipient_name}'s cards in the current round step by step.\n"
 " Potential {recipient_name}'s current believes about your cards: Understanding all given information and your knowledge about the {game_name}, if you were {recipient_name} (he can only observe your actions but cannot see your cards), please infer the {recipient_name}'s believes about your cards with probability (normalize to number 100% in total) step by step. Output: {agent_name} did action1 (probability) (after I did action or not) in the 1st round, , ... continue...  {agent_name} did action1 (probability) (after I did action or not)  in the current round, from the perspective of {recipient_name}, {agent_name} tends to have card1 (probability), card2 (probability) ... (normalize to number 100% in total).\n"
" Guess potential {recipient_name}'s goals in the next round: Understanding all given information and your knowledge about the {game_name}, if you were {recipient_name} (you can only observe his actions but cannot see his cards), please infer the {recipient_name}'s goal (if possible) after {agent_name} take the action. Output: when I do action1 (probability), {recipient_name} will do ...continue...\n"
 "Please don't respond too much irrelevant information."
\end{lstlisting}
Self-Aware Belief Prompt:
\begin{lstlisting}
"You are the player behind a NPC character called {agent_name}, and you are playing the board game {game_name} against {recipient_name}. \n"
" The game rule is: {rule} \n"
" Your estimated judgement about your own behaviour reflextion and improved strategy is: {pattern} \n"
" Your estimated judgement about the behaviour reflextion and improved strategy of {recipient_name} is: {oppo_pattern} \n"
" Your estimated belief about {recipient_name} is: {oppo_belief} \n"
" Your observation now is: {observation}\n"
 " Your current game progress summarization including actions and conversations with {recipient_name} is: {recent_observations}\n"
 " Your previous game memory including observations, actions, conversations with {recipient_name} and your reflection is: \n{long_memory}\n"
" Understanding the game rule, the cards you have, your observation,  progress summarization in the current game and previous game history, your estimated action reflection and improved strategy, and your knowledge about the {game_name}, can you do following things? \n"
 " Analyze your cards: Understanding all given information and your knowledge about the {game_name}, please analysis what is your possible combination, advantages and disadvantages of your cards in the current round step by step.\n"
" Make your goals of the current round: in order to win more chips in the end, think what words you want to say and what action you take to bluff {recipient_name}.\n "
"Please don't respond too much irrelevant information."
\end{lstlisting}
\subsection{Plan Recommendation}
\begin{lstlisting}
"You are the objective player behind a NPC character called {initiator_name}, and you are playing the board game {game_name} against {recipient_name}.\n"
" The game rule is: {rule} \n"
'{pattern}\n'
" Your observation about the game status now is: {observation}\n"
" Your current game progress summarization including actions and conversations with {recipient_name} is: {recent_observations}\n"
'{belief}\n'
" Understanding all given information, can you do following things:"
" Make Reasonable Plans: Please plan several strategies according to actions {valid_action_list} you can play now to win the finally whole {game_name} games step by step. Note that you can say something or keep silent to confuse your opponent.\n"
" List potential {recipient_name}'s actions and Estimate Winning/Lose/Draw Rate for Each Plan: From the perspective of {recipient_name} , please infer what action {recipient_name} wiil take with probability (normalize to number 100% in total) would do when {recipient_name} holds different cards and then calculate the winning/lose/draw rates when {recipient_name} holds different cards step by step. At last, please calculate the overall winning/lose/draw rates for each plan step by step considering  {recipient_name}'s behaviour pattern. Output in a tree-structure: "
"Output: Plan 1:  If I execute plan1.  "
"The winning/lose/draw rates when {recipient_name} holds card1: Based on {recipient_name}'s behaviour pattern, In the xx round, because {recipient_name} holds card1  (probability) and the combination with current public card (if release)  (based on my belief on {recipient_name}), and if he sees my action, {recipient_name} will do action1 (probability) ( I actually hold card and the public card (if reveal) is , he holds card1 and the public card (if reveal), considering Single Game Win/Draw/Lose Rule, please infer I will win/draw/lose  step by step ), action2 (probability) (considering Single Game Win/Draw/Lose Rule, please infer I will win/draw/lose step by step  ),.. (normalize to number 100% in total); \n   Overall (winning rate for his card1) is (probability = his card probability * win action probability), (lose rate for his card2) is (probability= his card probability * lose action probability), (draw rate for his card2) is (probability = his card probability * draw action probability)  "
"The winning/lose/draw rates when {recipient_name} holds card2: Based on {recipient_name}'s behaviour pattern, In the xx round, because {recipient_name} holds card2  (probability) and the combination with current public card (if release)  (based on my belief on {recipient_name}) , and if he sees my action, he will do action1 (probability) (I actually hold card and the public card (if reveal) is , he holds card1 and the public card (if reveal), considering Single Game Win/Draw/Lose Rule, please infer I will win/draw/lose  step by step ).. action2 (probability) (normalize to number 100% in total) (considering Single Game Win/Draw/Lose Rule, please infer I will win/draw/lose step by step ),.. ;..... continue ....\n Overall (winning rate for his card2) is (probability = his card probability * win action probability), (lose rate for his card2) is (probability= his card probability * lose action probability), (draw rate for his card2) is (probability = his card probability * draw action probability) "
"Plan1 overall {initiator_name}'s Winning/Lose/Draw rates : the Winning rate (probability) for plan 1 is (winning rate for his card1) + (winning rate for his card2) + .. ; Lose rate (probability) for plan 1 : (lose rate for his card1) + (lose rate for his card2) + .. ; Draw Rate (probability) for plan 1  : (draw rate for his card1) + (draw rate for his card2) + ... ;  (normalize to number 100% in total) for plan1 \n"
"Plan 2: If I execute plan2, The winning/lose/draw rates when {recipient_name} holds card1: Based on {recipient_name}'s behaviour pattern, In the xx round, if {recipient_name} holds card1  (probability)  and the combination with current public card (if release),  .. (format is similar with before ) ... continue .."
"Plan 3: .. Coninue ... "
" The number of payoffs for each plan: Understanding your current observation,  each new plans, please infer the number of wininng/lose payoffs for each plan step by step, Output: Plan1: After the action, All chips are in the pot:  If win, the winning payoff would be (Calculated by Winning Payoff Rules step by step) . After the action,  All chips in the pot:  If lose , the lose payoff would be:  (Calculated by Lose Payoff Rules step by step). Plan2:  After the action, All chips in the pot:  If win, the winning chips would be (Calculated by Winning Payoff Rules step by step):  After the action, All chips in the pot:  If lose , the lose chips would be:  (Calculated by Lose Payoff Rules step by step). If the number of my chips in pots have no change, please directly output them. \n"
" Estimate Expected Chips Gain for Each Plan : Understanding all the information and Estimate Winning/Lose/Draw Rate for Each Plan, please estimate the overall average Expected Chips Gain for each plan/strategy in the current game by calculating winning rate * (Winning Payoff Rule in the game rule) - lose rate * (Lose Payoff Rule in the game rule) step by step (Note that you should first consider whether to add chips into pot while taking differnt actions, then calculate the ecpected chip gain).\n"
" Plan Selection: Please output the rank of estimated expected chips gains for every plan objectively step by step, and select the plan/strategy with the highest estimated expected chips gain considering both the strategy improvement. \n "
\end{lstlisting}
\end{document}